\documentclass{article} 
\usepackage{iclr2016_conference,times}
\usepackage{url}
\usepackage{graphicx}
\usepackage{amsmath}
\usepackage{amssymb}
\usepackage{float}
\usepackage{subfig,caption}
\usepackage[colorlinks = true,
            linkcolor = blue,
            urlcolor  = blue,
            citecolor = blue,
            anchorcolor = blue]{hyperref}

\title{Learning Visual Predictive Models of Physics for Playing Billiards}

\author{Katerina Fragkiadaki\thanks{equal contribution} 
\quad
Pulkit Agrawal$^*$
\quad
Sergey Levine
\quad
Jitendra Malik \\
Electrical Engineering and Computer Science\\
University of California, Berkeley\\
\texttt{(katef,pulkitag,svlevine,malik)@berkeley.edu} \\
}

\begin{document}

\maketitle

\newcommand{\mocap}{\theta}
\newcommand{\ea}{\text{\textit{et. al}}}
\newcommand{\z}{\mathrm{z}}
\newcommand{\tr}{\mathrm{tr}}
\newcommand{\ntr}{n_T}
\newcommand{\auto}{\text{ERD}}
\newcommand{\ERD}{\text{ERD}}
\newcommand{\data}{x}
\newcommand{\deltax}{\Delta x}
\newcommand{\deltay}{\Delta y}
\newcommand{\horizon}{h}
\newcommand{\seq}{\mathbf{x}}
\newcommand{\WW}{\mathbf{W}}
\newcommand{\xex}{\mathbf{x}}
\newcommand{\Lo}{\mathcal{L}}
\newcommand{\thetanet}{\boldsymbol \theta}
\newcommand{\LL}{\mathbf{L}}
\newcommand{\clcl}{\mathrm{cl}^2}
\newcommand{\dl}{\mathrm{dl}}
\newcommand{\ov}{s}
\newcommand{\ATsteer}{\W^{\mathrm{steer}}_{T}}

\newcommand{\Csteer}{\C^{\mathrm{steer}}}
\newcommand{\ndl}{n_D}
\newcommand{\np}{n_P}
\newcommand{\salgroup}{\mathrm{ncut}}
\newcommand{\cnfd}{\mathrm{confidence}}
\newcommand{\thresh}{\mathrm{th}}
\newcommand{\dr}{\mathrm{r}}
\newcommand{\dc}{\ce}
\newcommand{\confcf}{\mathbf{confidence}}
\newcommand{\aligncf}{\mathrm{align}}
\newcommand{\AD}{\mathbf{A}_D}
\newcommand{\W}{\mathbf{W}}
\newcommand{\eig}{\mathrm{eig}}
\newcommand{\pbth}{\beta}
\newcommand{\discthresh}{\gamma}
\newcommand{\Asteer}{\W^\mathrm{steer}_T}
\newcommand{\Asteernoh}{\mathbf{A}_\mathbf{steer}}
\newcommand{\Aco}{\mathbf{A}^{\mathrm{c}}}
\newcommand{\RD}{\mathbf{R}_D}
\newcommand{\dlsel}{h}
\newcommand{\RDF}{\mathbf{R}_D^\dlsel}
\newcommand{\R}{\mathcal{R}}
\newcommand{\AT}{\mathbf{A}}
\newcommand{\f}{\mathrm{f}}
\newcommand{\trc}{\mathrm{C}}
\newcommand{\RT}{\mathbf{R}_T}
\newcommand{\Tf}{\mathcal{T}^\mathbf{f}}
\newcommand{\st}{\mathrm{subject \text{ } to}}
\newcommand{\vc}{\mathrm{vec}}
\newcommand{\Wtilde}{\tilde{\mathbf{W}}}
\newcommand{\thr}{\mathrm{th}}
\newcommand{\maxmarg}{\mathrm{mxmr}}
\newcommand{\trd}{\tr^D}
\newcommand{\C}{\mathbf{C}}
\newcommand{\ns}{n_S}
\newcommand{\nss}{m}
\newcommand{\Cg}{\mathbf{C}_G}
\newcommand{\Prob}{\mathrm{P}}
\newcommand{\D}{\mathbf{D}}

\newcommand{\g}{\mathrm{G}}

\newcommand{\score}{c}

\newcommand{\bbox}{\mathrm{box}}
\newcommand{\X}{\mathrm{X}}
\newcommand{\T}{\mathcal{T}}
\newcommand{\diag}{\mathrm{Diag}}
\newcommand{\Dcap}{\mathcal{D}}
\definecolor{lightgray}{gray}{0.9}
\newcommand{\todo}{\textcolor{red}{TODO: }}
\newcommand{\velocity}{\mathrm{vel}}
\newcommand{\disparity}{\mathrm{dsp}}
\newcommand{\trajdist}{\mathrm{dst}}
\newcommand{\Gup}{\mathrm{G}}
\newcommand{\detection}{R}

\newcommand{\na}{n(\allX)}
\newcommand{\location}{\mathrm {loc}}

\newcommand{\Tr}{\mathrm {Tr}}
\newcommand{\YY}{\mathcal{Y}}
\newcommand{\sel}{\mathbf {s}}
\newcommand{\yf}{y}
\newcommand{\ATDh}{\mathbf{A}_T^{\yf}}
\newcommand{\Ch}{\mathrm{steer}(\C;h)}
\newcommand{\RTh}{\mathbf{R}_T^{\yf}}
\newcommand{\AR}{\mathbf{A}_R}
\newcommand{\pn}{\mathrm{Partition}}
\newcommand{\ptwisemul}{\bullet}

\newcommand{\modify}{M}
\newcommand{\zeroone}{\{0, 1\}}
\newcommand{\ones}{\mathbf 1}
\newcommand{\suchthat}{\mathrm{s.t.}}
\newcommand{\A}{\mathbf{A}}
\newcommand{\GF}{\mathbb{G}}
\newcommand{\e}{e}
\newcommand{\nmultseg}{|\mathrm{Seg}|}
\newcommand{\cut}{\mathrm{cut}}
\newcommand{\degree}{\mathrm{degree}}
\newcommand{\npool}{n(\allX)}
\newcommand{\gscal}{\pbth}
\newcommand{\Dr}{\mathbf{D}}
\newcommand{\allX}{\mathbf X}
\newcommand{\allY}{\mathbf Y}
\newcommand{\footnoteremember}[2]{
\footnote{#2}
\newcounter{#1}
\setcounter{#1}{\value{footnote}}
}
\newcommand{\footnoterecall}[1]{
}

\newcommand{\hh}{h}
\newcommand{\p}{\mathbf{p}}
\newcommand{\link}{\mathrm{links}}
\newcommand{\vol}{\mathrm{vol}}
\newcommand{\ncut}{\mathrm{ncut}}
\newcommand{\AF}{\mathbb{A}}
\newcommand{\BF}{\mathbb{B}}
\newcommand{\VF}{\mathbb{V}}
\newcommand{\px}{\mathrm{p}}
\newcommand{\Edge}{E^D}
\newcommand{\nr}{n_R}
\newcommand{\xx}{\mathbf{x}}
\newcommand{\ww}{\mathbf{w}}
\newcommand{\wwp}{\mathbf{w}^{D}}

\newcommand{\aex}{a}
\newcommand{\ca}{c}
 \newcommand{\nd}{n_D}

 \newcommand{\Wbar}{\mathbf{\hat{W}}}
 \newcommand{\RSPX}{\mathbf{R}_R}
%

%
%
\newcommand{\DD}{\mathcal D}

%
\newcommand{\nS}{n_{S}}
\newcommand{\den}{\mathrm{\rho}}
\newcommand{\spx}{\mathrm{r}}
\newcommand{\prt}{\mathrm{d}}
\newcommand{\Seg}{\mathbf{S}}
\newcommand{\aff}{\ww}
\newcommand{\affr}{\ww^R}
\newcommand{\Diag}{\mathrm{Diag}}
\newcommand{\conf}{\mathbf{c}}
\newcommand{\stab}{\mathbf{ncut}}
\newcommand{\cmtb}{\mathbf{cmtb}}
\newcommand{\pset}{\mathcal{P}}

\newcommand{\attr}{\mathrm{att}}

%
%
\newcommand{\V}{\mathbf{V}}
\newcommand{\Y}{\mathrm{Y}}

\newcommand{\rr}{\mathrm{r}}
\newcommand{\pa}{p}
\newcommand{\EV}{\mathbf{\Lambda}}
\newcommand{\WE}{\mathbf{W}}
\newcommand{\K}{K}
\newcommand{\ce}{\mathbf{c}}
\newcommand{\width}{\mathrm{w}}
\newcommand{\height}{\mathrm{h}}
\newcommand{\nf}{n_F}
\newcommand{\sss}{\tilde{s}}
\newcommand{\PS}{\mathbf{P}}
\newcommand{\SSp}{\mathcal{S}}
\newcommand{\N}{\mathcal{N}}
\newcommand{\DT}{G}
\newcommand{\di}{\mathbf{d}}
\newcommand{\edge}{\mathrm{e}}
 \newcommand{\ASPX}{\mathbf{A}_{R}}
 \newcommand{\ASPXst}{\mathbf{W}^{\mathrm{steer}}}
 \newcommand{\PP}{\mathcal{D}}
 \newcommand{\bo}{t_f}
%
%

%
 \newcommand{\s}{\mathcal S (\mathcal D)}

\begin{abstract}

The ability to plan and execute goal specific actions in varied and unseen environments settings is a central requirement of intelligent agents. In this paper, we explore how an agent can be equipped with an internal model of the dynamics of the external world, and how it can use this model to plan novel actions by running multiple internal simulations (``visual imagination''). Our models directly process raw visual input, and use a novel object-centric prediction formulation based on visual glimpses centered on objects (fixations) to enforce translational invariance of the learned physical laws. The agent trains itself through random interaction with a collection of different environments, and the resulting model can then be used to plan goal-directed actions in novel environments that were previously never encountered by the agent. We demonstrate that our agent can accurately plan actions for playing a simulated billiards game, which requires pushing a ball into a target position or into collision with another ball.

\end{abstract}

\section{Introduction}
Imagine a hypothetical person who has never encountered the game of billiards. While this person may not be very adept at playing the game, he would still be capable of inferring the direction in which the cue ball needs to be hit to displace the target ball to a desired location. How can this person make such an inference without any prior billiards-specific experience? One explanation is that humans are aware of the laws of physics, and a strategy for playing billiards can be inferred from knowledge about dynamics of bouncing objects. However, humans do not appear to consciously solve Newton's equations of motion, but rather have an intuitive understanding of how their actions affect the world. In the specific example of billiards, humans can imagine the trajectory that the ball would follow when a force is applied, and how the trajectory of ball would change when it hits the side of the billiards table or another ball. We term models that can enable the agents to visually anticipate the future states of the world as visual predictive models of physics. 

A visual predictive model of physics equips an agent with the ability to generate potential future states of the world in response to an action without actually performing that action (``visual imagination''). Such visual imagination can be thought of as running an internal simulation of the external world. By running multiple internal simulations to imagine the effects of different actions, the agent can perform planning, choosing the action with the best outcome and executing it in the real world. The idea of using internal models for planning actions is well known in the control literature \citep{Mayne20142967}. However, the question of how such models can be learned from raw visual input has received comparatively little attention, particularly in situations where the external world can change significantly, requiring generalization to a variety of environments and situations.

Previous methods have addressed the question of learning models, including visual models, of the agent's own body \citep{DBLP:journals/corr/WatterSBR15,DeepMindContinuous}. However, when performing actions in complex environments, models of both the agent and the external world are required. The external world can exhibit considerably more variation than the agent itself, and therefore such models must generalize more broadly. This makes problem of modelling the environment substantially harder than modelling the agent itself. 

The complexities associated with modeling the external world may be elucidated through an example. Consider the family of worlds composed of moving balls on a 2D table (i.e. \textit{moving-ball} world). This family contains diverse worlds that can be generated by varying factors such as the number of balls, the table geometries, ball sizes, the colors of balls and walls, and the forces applied to push the balls. Because the number of objects can change across worlds, it is not possible to explicitly define a single state space for these worlds. For the purpose of modeling, an implicit state space must be learnt directly from visual inputs. In addition to this combinatorial structure, differences in geometry and nonlinear phenomena such as collisions result in considerable complexity.

Similar to the real world, in the \textit{moving-ball} world, an agent must perform actions in novel conditions it has never encountered before. Although \textit{moving-ball} worlds are relatively constrained and synthetic, the diversity of such worlds can be manipulated using a small number of factors. This makes them a good case study for systematically evaluating and comparing the performance of different model-based action selection methods under variation in external conditions (i.e. generalization).

\begin{figure*}
\begin{center}
\includegraphics[scale=0.29]{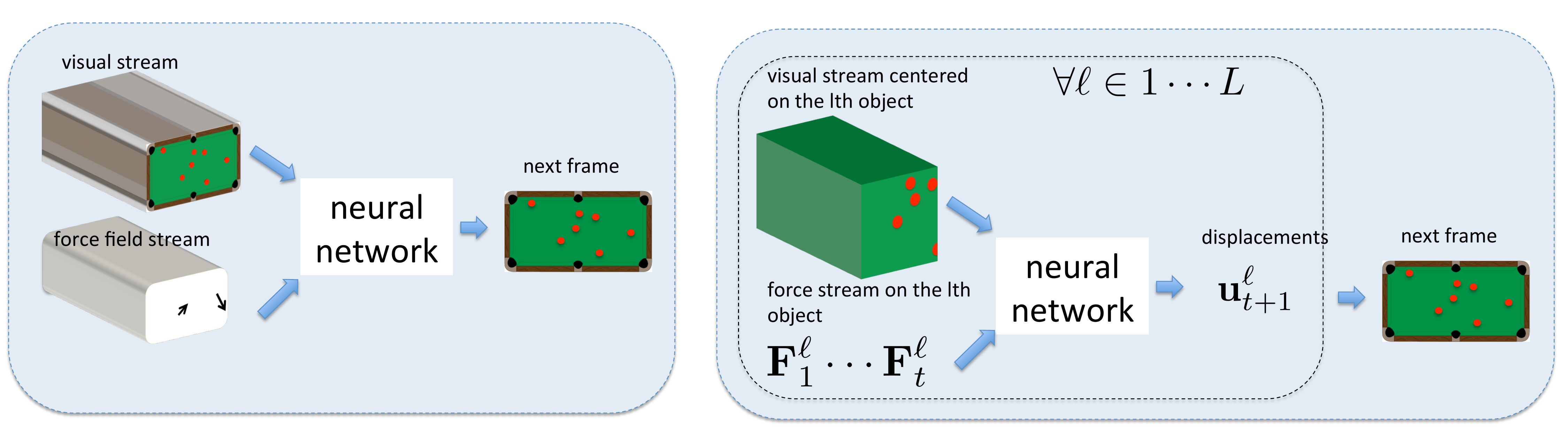}
\end{center}
\caption{ \textbf{Frame-centric versus object-centric prediction. } \textit{Left:} Frame-centric model predicts takes as input the the image of the the entire billiards and forces applied by the agent to make predictions about the future. \textit{Right:} Object-centric model predicts the future states of the world by individually modeling the temporal evolution of each the L objects in the world. In the billiards world, predicting the future velocities $(u_{t+h}^l, h \in [1, 20], l \in [1, L])$ of each of the L balls is sufficient for generating the future world states. Please see section \ref{sec:learning} for more details. 
}
\label{fig:objectcentric}
\end{figure*}

Both the real world and the synthetic \textit{moving-ball} worlds also contain regularities that allow learning generalizable models in the face of extensive variation, such as the translational invariance of physical laws. The main contribution of this work is a first step towards learning dynamical model of the external world directly from visual inputs that can handle combinatorial structure and exploits translation invariance in dynamics. We propose an object-centric (OC) prediction approach, illustrated in Figure \ref{fig:objectcentric}), that predicts the future states of the world by individually modeling the temporal evolution of each object from object-centric glimpses. The object-centric (OC) approach naturally incorporates translation invariance and model sharing across different worlds and object instances.





We use a simulated billiards-playing domain where the agent can push balls on a 2D billiard table with varying geometry as a working example to evaluate our approach. We show that our agent learns a model of the billiards world that can be used to effectively simulate the movements of balls and consequently plan actions without requiring any goal-specific supervision. Our agent successfully predicts forces required to displace the ball to a desired location on the billiards table and to hit another moving ball. 

\section{Previous Work}





\paragraph{Vision-based control}

Due to the recent success of deep neural networks for learning feature representations that can handle the complexity of visual input \cite{NIPS2012_0534}, there has been considerable interest in utilizing this capability for learning to control dynamical systems directly from visual input. Methods that directly learn policies for prediction actions from visual inputs have been successfully used to learn to play Atari games \cite{DBLP:journals/corr/MnihKSGAWR13} and control a real robot for a predefined set of manipulation tasks \cite{DBLP:journals/corr/LevineFDA15}. However, these methods do not attempt to model how visual observations will evolve in response to agent's actions. This makes it difficult to repurpose the learned policies for new tasks.

Another body of work \citep{DBLP:conf/icmla/KietzmannR09,DBLP:conf/ijcnn/LangeRV12} has attempted to build models that transform raw sensory observations into a low-dimensional feature space that is better suited for reinforcement learning and control. More recently works such as  \citep{DBLP:journals/corr/WahlstromSD15,DBLP:journals/corr/WatterSBR15} have shown successful results on relatively simple domains of manipulating a synthetic two degree of freedom robotic arm or controlling an inverted pendulum in simulation. However, training and testing environments in these works were exactly the same. In contrast, our work shows that vision based model predictive control can be used in scenarios where the test environments are substantially different from training environments.

\paragraph{Models of physics and model based control}
\citep{hamrick2011internal} provided evidence that human judgement of how dynamical systems evolve in future can be explained by the hypothesis that humans use internal models of physics. \citep{jordan1992forward,wolpert1995internal, haruno2001mosaic, todorov2003unsupervised} proposed using internal models of the external world for planning actions. However these works have either been theoretical or have striven to explain sensorimotor learning in humans. To the best of our knowledge we are the first work that strives to build an internal model of the external world purely from visual data and use it for planning novel actions. \citep{actionconditioned} successfully predict future frames in Atari game videos and train a Q-controller for playing Atari games using the predicted frames. Training a Q-controller requires task specific supervision whereas our work explores whether effective dynamical models for action planning can be learnt without requiring any task specific supervision. 


\paragraph{Learning Physics from Images and Videos}
Works of \cite{NIPS2015_5780,conf/eccv/BhatSP02,5459407,DBLP:journals/corr/MottaghiBRF15} propose methods for estimating the parameters of Newtonian equations from images and videos. As laws of physics governing the dynamics of balls and walls on a billiards table are well understood, it is possible to use these laws instead of learning a predictive model for planning actions. However, there are different dynamic models that control ball-ball collisions, ball-wall collisions and the movement of ball in free space. Therefore, if these known dynamical model are to be used, then a system for detecting different event types would be required for selecting the appropriate dynamics model at different time points. In contrast, our approach avoids hand designing such event detectors and switches and provides a more general and scalable solution even in the case of billiards.

\paragraph{Video prediction}
\citep{DBLP:conf/nips/MichalskiMK14,sht-rtrbm-08} learn models capable of generating images of bouncing balls. However, these models are not shown to generalize to novel environments. Further these works donot include any notion of an agent or its influence on the environment. \citep{DBLP:conf/icra/BootsBF14} proposes model for predicting the future visual appearance of a robotic arm, but the method is only shown to work when the same object in the same visual environment is considered. Further it is not obvious how the non-parametric approach would scale with large datasets. In contrast our approach generalized to novel environments and can scale easily with large amounts of data. 

\paragraph{Motion prediction  for Visual Tracking}
In Computer Vision, object trackers use a wide variety of predictive models, from simple constant velocity models, to linear dynamical systems and their variants \citep{Urtasun:2006:PTG:1153170.1153448}, HMMs \citep{Brand:1997:CHM:794189.794420,Ghahramani:1997:FHM:274158.274165}, and other models. Standard smoothers or filters, such as Kalman filters \citep{Weng:2006:VOT:1223195.1223208}, usually operate on Cartesian coordinates, rather than the visual content of the targets, and in this way discard useful information that may be present in the images. Finally, methods for  3D tracking of  \cite{DBLP:conf/bmvc/KyriazisOA11,conf/iccv/SalzmannU11}  use Physics simulators to constrain the search space during data association.  
\section{Learning Predictive Visual Models}
\label{sec:learning}
We consider an agent observing an interacting with dynamic \textit{moving-ball} worlds consisting of multiple balls and multiple walls. We also refer to these worlds as billiard worlds. The agent interacts with the world by applying forces to change the velocities of the balls. In the real world, the environment of an agent is not fixed, and the agent can find itself in environments that it has not seen before. To explore this kind of generalization, we train our predictive model in a variety of billiards environments, which involve different numbers of balls and different wall geometries, and then test the learnt model in previously unseen settings.

In the case of \textit{moving-ball} world, it is sufficient to predict the displacement of the ball during the next time step to generate the visual of the world in the future. Therefore, instead of directly predicting image pixels, we predict each object's current and future velocity given a sequence of visual glimpses centered at the object (visual fixation) and the forces applied to it. 

We assume that during training the agent can perfectly track the objects. This assumption is a mild one because not only tracking is a well studied problem but also because there is evidence in the child development literature that very young infants can redirect their gaze to keep an object in focus by anticipating its motion (i.e. smooth pursuit) \citep{Hofsten19971799}. The early development of smooth pursuit suggests that it is important for further development of visual and motor faculties of a human infant. 


\begin{figure*}[t]
\begin{center}
\includegraphics[scale=0.38]{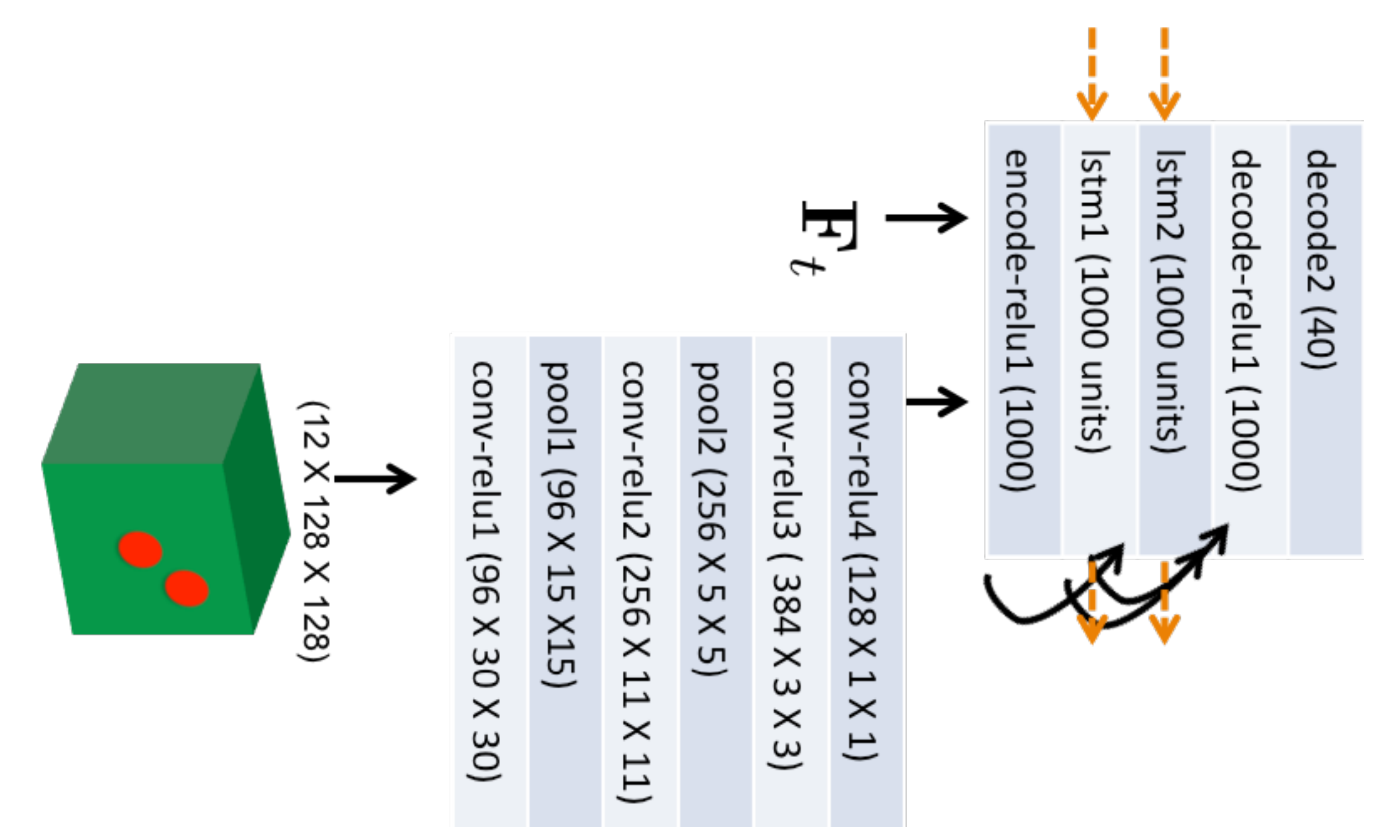}
\end{center}
\caption{ \textbf{Network architecture.} At each time step $t$, for each object, the network is provided with the previous four glimpses centered on the object's position, as well as the agent's applied forces $\mathbf{F}_t=(F_t^x, F_t^y)$ and the hidden states of the LSTM units from the previous time step.
The output is ball displacements $\mathbf{u}_{t+k} =(\deltax_{t+k},\deltay_{t+k})$ for $k=1 \cdots \horizon$ in the next $\horizon$ frames.
}
\label{fig:architecture}
\end{figure*}

\subsection{Model Details}
Our network architecture is illustrated in figure \ref{fig:architecture}. The input to the model is a stack of 4 images comprised of the current and previous 3 glimpses of the fixated object and the exercised force on the object at the current time step. The model predicts the velocity of the object at each of the $\horizon$ time steps in the future. We use $\horizon$ = 20. The same model is applied to all the objects in the world. 

Our network uses an AlexNet style architecture \citep{NIPS2012_0534} to extract visual features. The first layer (conv1) is adapted to process a stack of 4 frames. Layers 2 and 3 have the same architecture as that of AlexNet. Layer 4 (conv4) is composed of 128 convolution kernels of size $3\times3$. The output of conv4 is rescaled to match the value range of the applied forces, then is concatenated with the current force  and is passed into a fully connected (encoder) layer. 
Two layers of LSTM units operate on the output of the encoder to model long range temporal dynamics. Then, the output is decoded to  predicted velocities.



The model is trained by minimizing the Euclidean loss between ground-truth and predicted object velocities for $\horizon$ time steps in the future. The ground-truth velocities are known because we assume object tracking. The loss is mathematically expressed as:  
\begin{equation}
\mathcal{L}=\displaystyle\sum_{k=1}^\horizon w_k \|\tilde{\mathbf{u}}_{t+k}-\mathbf{u}_{t+k}\|_2^2
\end{equation}
where $\mathbf{u}_{t+k}, \tilde{\mathbf{u}}_{t+k}$ represent ground-truth and predicted velocities at the $k^{th}$ time step in the future respectively. Velocities are expressed in cartesian coordinates. We weigh the loss in a manner that errors in predictions at a shorter time horizon are penalized more than predictions at a longer time horizon. This weighting is achieved using penalty weights $w_k=\exp(-k^{\frac{1}{4}})$. We use the publicly available Caffe package for training our model. 

\begin{figure*}[t!]
\begin{center}
\includegraphics[trim=0.0in 0in 0in 0in, scale=0.36]{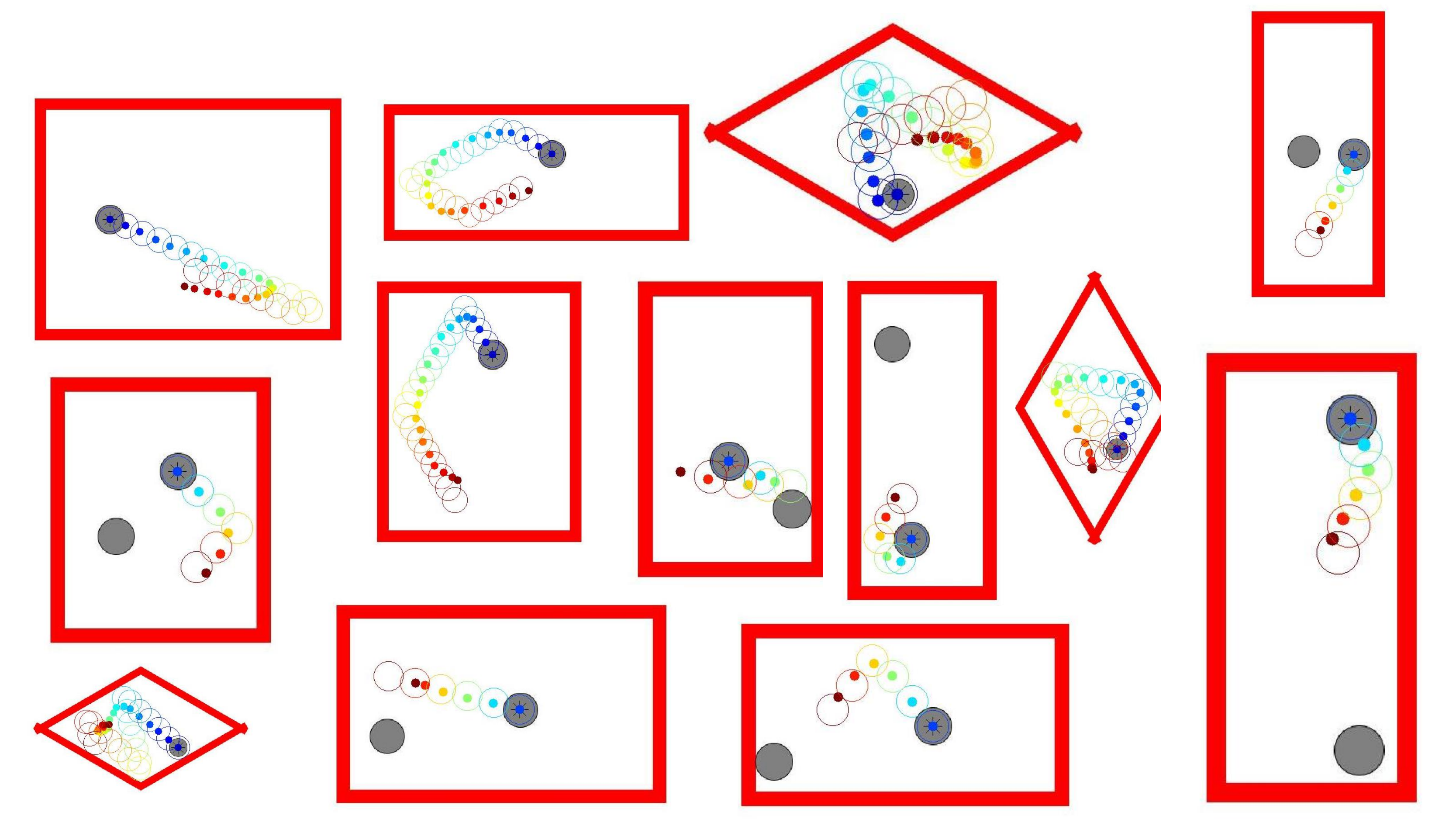}
\end{center}
\caption{Predicting ball motion under interactions and agent forces. Dots denote predictions of our model and  circles the ground-truth future positions for the fixated ball. Color indicates length of the prediction horizon,  blue denotes  close and red far in the future. Our model accurately predicts collisions under a wide variety of table configurations.}
\label{fig:trajectories}
\end{figure*}

For model learning, we generate sequences of ball motions in a randomly sampled world configuration. As shown in Figure \ref{fig:trajectories}, we experimented both with rectangular and non-rectangular wall geometries. For rectangular walls, a single sample of the world was generated by randomly choosing the size of the walls, location of the balls and the forces applied on the balls from a predefined range. The length of each sequence was sampled from the range [20, 200]. The length of the walls was sampled from a range of [300 pixels, 550 pixels]. Balls were of radius 25 pixels and uniform density. Force direction was uniformly sampled and the force magnitude was sampled from the range [30K Newtons, 80K Newtons]. Forces were only applied on the first frame. The size of visual glimpses is 600x600 pixels. The objects can move up to 10 pixels in each time step and therefore in 20 time steps they can cover distances up to 200 pixels. 

For training, we pre-generated 10K such sequences. We constructed minibatches by choosing 50 random subsets of 20 consequent frames from this pre-generated dataset. Weights in layers conv2 and conv3 were initialized from the weights of Alexnet that was trained for performing image classification on Imagenet \citep{NIPS2012_0534}. Weights in other layers are randomly initialized.


\section{Model Evaluation}
\label{sec:eval}
First we report evaluations on random worlds sampled from the same distribution as the training data. Next, we report evaluations on worlds sampled from a different distribution of world configurations to study the generalization of the proposed approach. Error in the angle and magnitude of the predicted velocities were used as performance metrics. We compared the performance of the proposed object centric (OC) model with a constant velocity (CV) and frame centric (FC) model. 
The constant velocity model predicts the velocity of a ball for all the future frames to be same as the ground truth velocity at the previous time step. The ball changes the velocity only when it strikes another ball or hits a wall. As collisions are relatively infrequent, the constant velocity model is a good baseline.   

We first trained a model on the family of rectangular worlds consisting of 1 ball only. The results of this evaluation are reported in Table \ref{tab:fc-oc}. We used average error across all the frames and the error averaged across frames only near the collisions as the error metrics for measuring performance. As balls move in linear trajectories except for time of collision, accurately predicting the velocities after a collision event is of specific interest. Results in Table \ref{tab:fc-oc} show that the object centric (OC) model is better than frame centric model (FC) model and much better than the constant velocity model. These results show that object centric modelling leads to better learning. 


\begin{center}
\begin{table}
\begin{center}
  \begin{tabular}{ l | c | c| c | c | c | c }
    Time & \multicolumn{3}{c|}{Overall Error} & \multicolumn{3}{c}{Error Near Collisions} \\
    \hline
    & CV & FC & OC & CV & FC & OC \\ \hline
    t+1  & $3.0^{o}$/0.00 &  $6.2^{o}$/0.04 & $5.1^{o}/0.03$   & $23.2^{o}$/0.00  & $11.4^{o}$/0.06 & $9.8^{o}$/0.04 \\ \hline
    t+5  & $11.8^{o}$/0.01 &  $8.7^{o}$/0.05 & $7.2^{o}/0.04$   & $56.6^{o}$/0.05  & $21.1^{o}$/0.12 & $17.9^{o}$/0.10\\ \hline
    t+20 & $45.3^{o}$/0.01 & $16.3^{o}$/0.09 & $14.8^{o}/0.09$ & $123.0^{o}$/0.04 & $54.8^{o}$/0.20 & $54.8^{o}$/0.20 \\ 
    \hline
  \end{tabular}
  \caption{Quantitative evaluation of the prediction model reported as error in the magnitude and angle of the velocity. The average error across all frames (i.e. Overall Error) and error averaged only in frames that were within [-4, 4] frames of a frame depicting collision (i.e. Error Near Collisions) are reported for Constant Velocity (CV) model, Frame Centric (FC) and Object Centric (OC) models. The errors are reported as $a^{o}/b$, where $a$ is the mean of angular error in degrees and $b$ is the relative error in the magnitude of the predicted velocity.  The constant velocity model predicts the velocity of a ball for all the future frames to be same as the ground truth velocity at the previous time step.}  
 \label{tab:fc-oc}
  \end{center}
\end{table}
\end{center}

\begin{figure}[t]
\begin{center}
\includegraphics[width=0.46\linewidth]{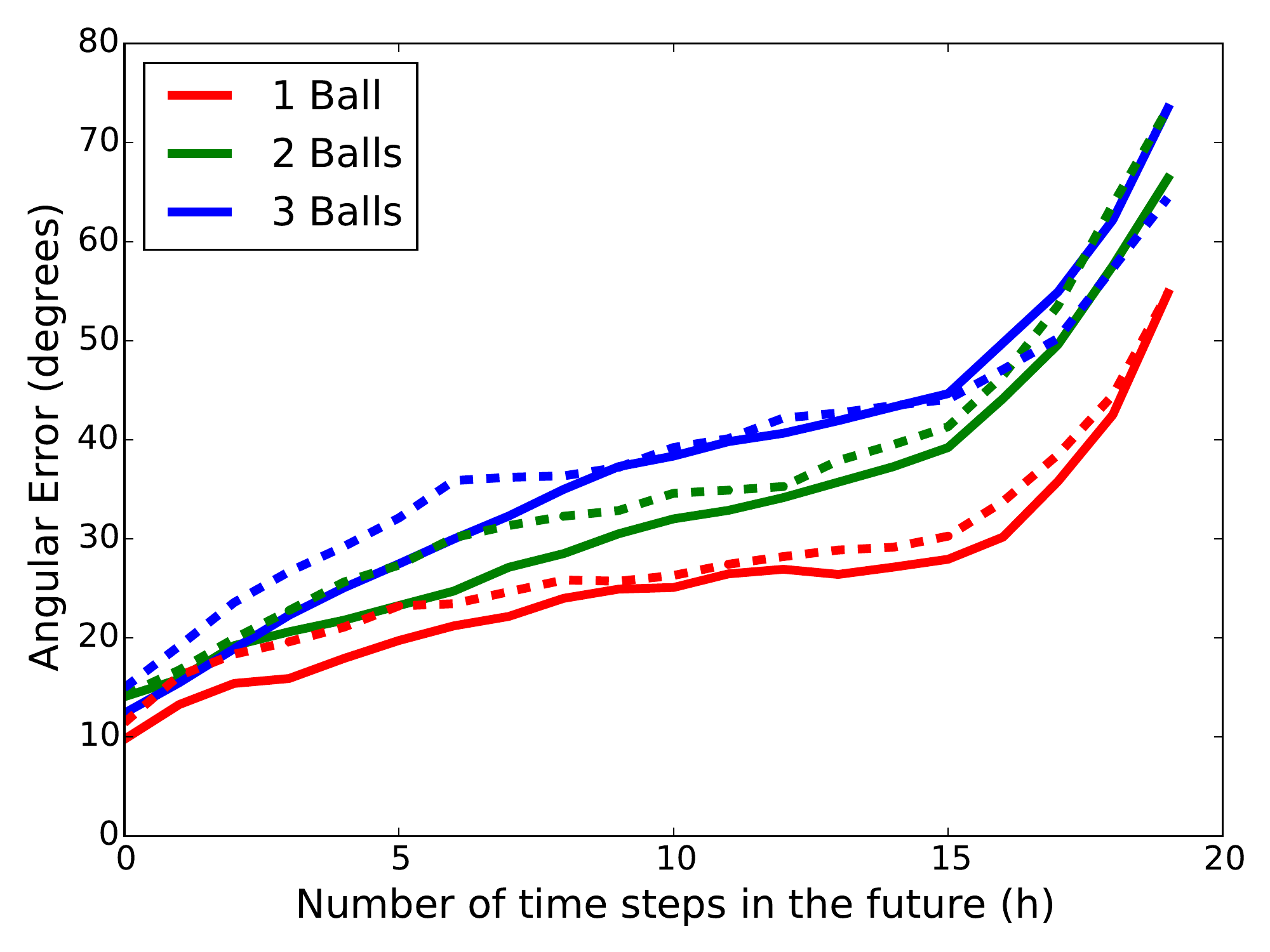}
\includegraphics[width=0.46\linewidth]{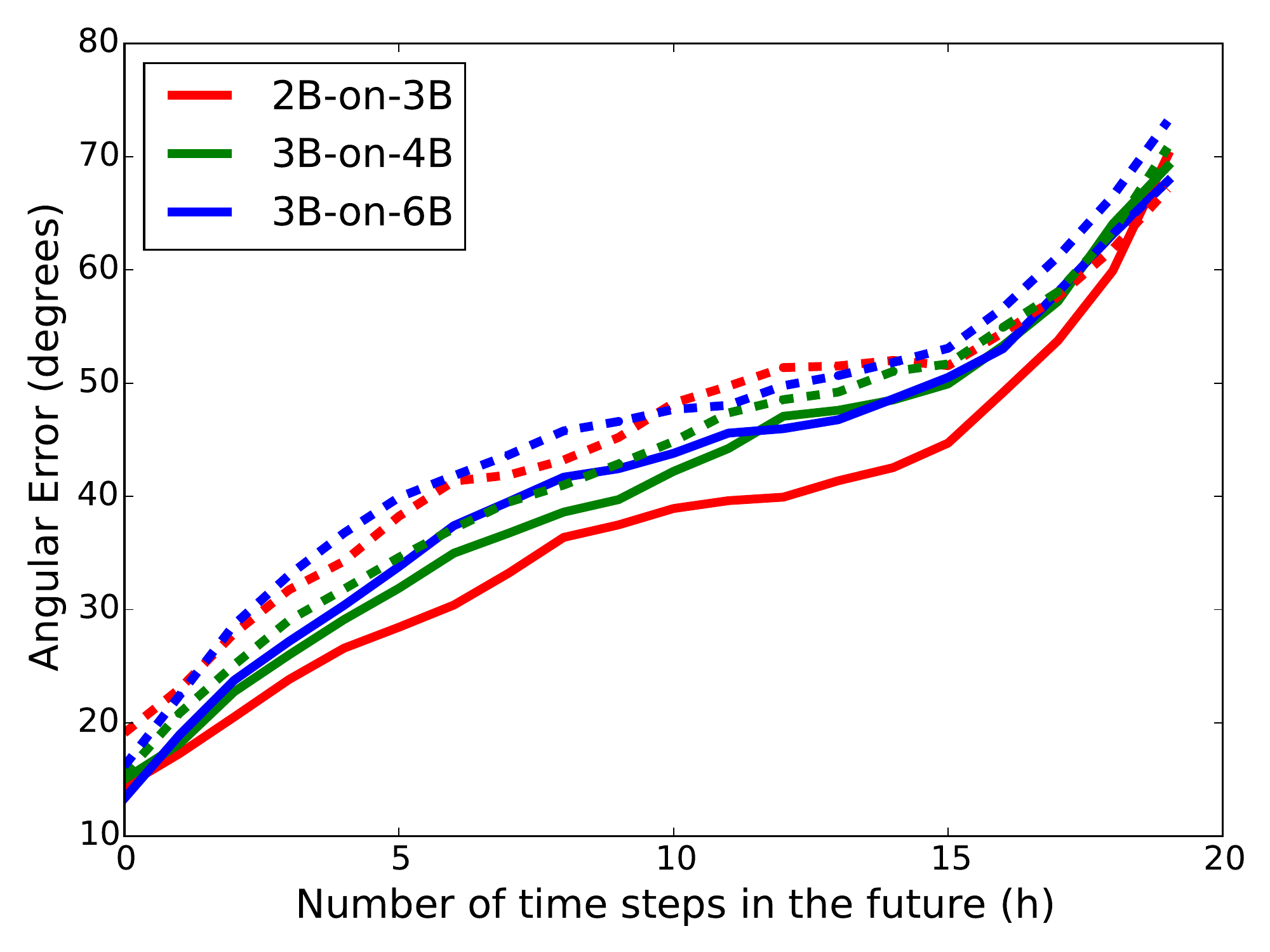}
\end{center}
\caption{Performance comparison of frame centric (FC) and object centric (OC) models. Performance is measured as the angular error near collisions. Dashed and solid lines show the angular error over a horizon of 20 time steps ($\horizon = 20$) for FC and OC models respectively. (\textit{Left}) Angular errors for models trained and tested on 1, 2 and 3 ball worlds respectively. The performance of the models only degrades slightly with increasing number of balls. Generally, the OC model is more accurate than the FC model. (\textit{Right}) Comparing the performance the FC model against the OC model when models trained on 2 and 3 balls are tested on worlds containing a larger number of balls. The naming convention nB-on-mB indicates models trained on n-ball worlds and tested on m-ball worlds. The generalization of the proposed OC model is significantly better than the FC model.}
\label{fig:nb-scaling}
\end{figure}

How well does our model scale with increasing number of balls? For studying this, we trained models on families of world consisting of 2 and 3 balls respectively. We used the learnt 1-ball model to initialize the training of the 2-ball model, which in turn was used to initialize the training of the 3-ball model. We found this curriculum learning approach to outperform models trained from scratch. The 2 and 3-ball models were evaluated on worlds separate from training set that consisted of 2 and 3 ball respectively. The angular errors measured near collisions (for $\horizon = $ 1 to 20) are shown Figure \ref{fig:nb-scaling}. The performance of our model degrades only by small amounts as the number of balls increase. Also, in general the OC model performs better than the FC model. 

We also trained and tested our models on non-rectangular walls. Qualitative visualizations of ground truth and predicted ball trajectories are show in figure \ref{fig:trajectories}. The figure shows that our model accurately predicts the velocities of balls after collisions in varied environments. This result indicates that our models are not constrained to any specific environment and have learnt something about the dynamics of balls and their collisions.

\subsection{Evaluating Generalization}

The results reported in the previous section show generalization to worlds sampled from the same distribution as the training set. In addition to this, we also tested our models on worlds substantially different from the worlds in the training set.

Figure \ref{fig:compare_lstm} shows that our model can generalize to much larger wall configurations than those used in the training. The wall lengths in the training set were between 300-550 pixels, whereas the the wall lengths in the testing set were sampled from the range of 800-1200 pixels. This shows that our models can generalize to different wall geometries. 

Figure \ref{fig:nb-scaling} (\textit{right}) shows that models trained on 2 and 3-ball worlds perform well when tested on 3, 4 and 6-ball worlds. This shows that our models can generalize to worlds with larger number of balls without requiring any additional training. The results in the figure also show that proposed OC model generalizes substantially better than the FC model.


\section{Generating Visual Imaginations}
\begin{figure}[t]
\begin{center}
\includegraphics[scale=0.24]{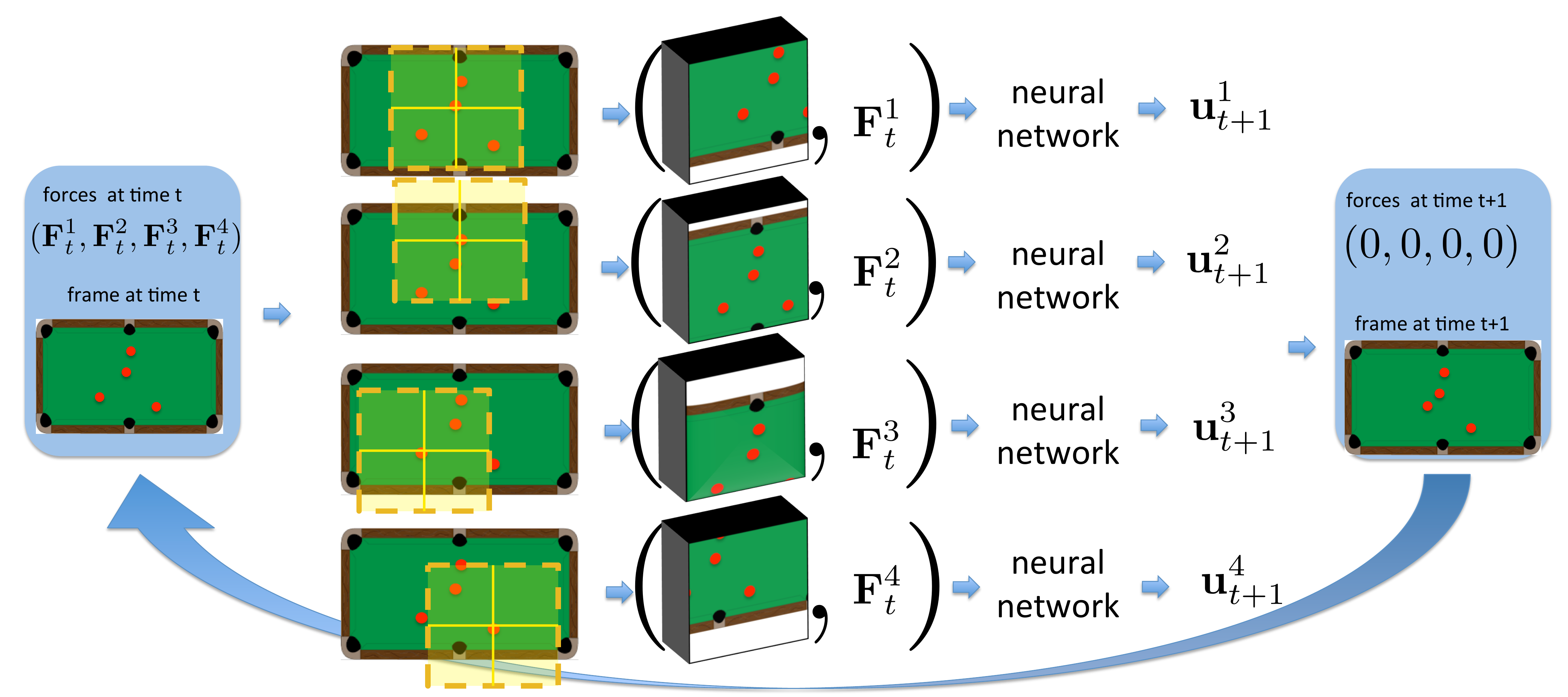}
\end{center}
\caption{Generating visual imaginations: The visual image of the world at current and past 3 time steps and the applied forces are used to predict the velocity of every ball. The visual image of the world at the next time step is rendered by translating the balls by their predicted velocities. This process is repeated iteratively to generate a sequence of visuals of the future world states.}
\label{fig:gen_hallucinations}
\end{figure}

As our models can accurately predict ball dynamics in the future, we can use these models to generate visual imaginations of the future. The procedure we use for generating visual imaginations is illustrated in figure \ref{fig:gen_hallucinations}. Given a configuration of the world and applied forces, the visual image of the world at the time step $t+1$ is generated by translating each ball by amount to equal to its predicted velocity ($\tilde u_t$) at time $t$. This generated image forms the input to the model for generating the visual image of the world at time step $t+2$. This process is repeated iteratively to generate visual imaginations of the future world states.

\begin{figure}[t]
\begin{center}
\includegraphics[scale=0.30]{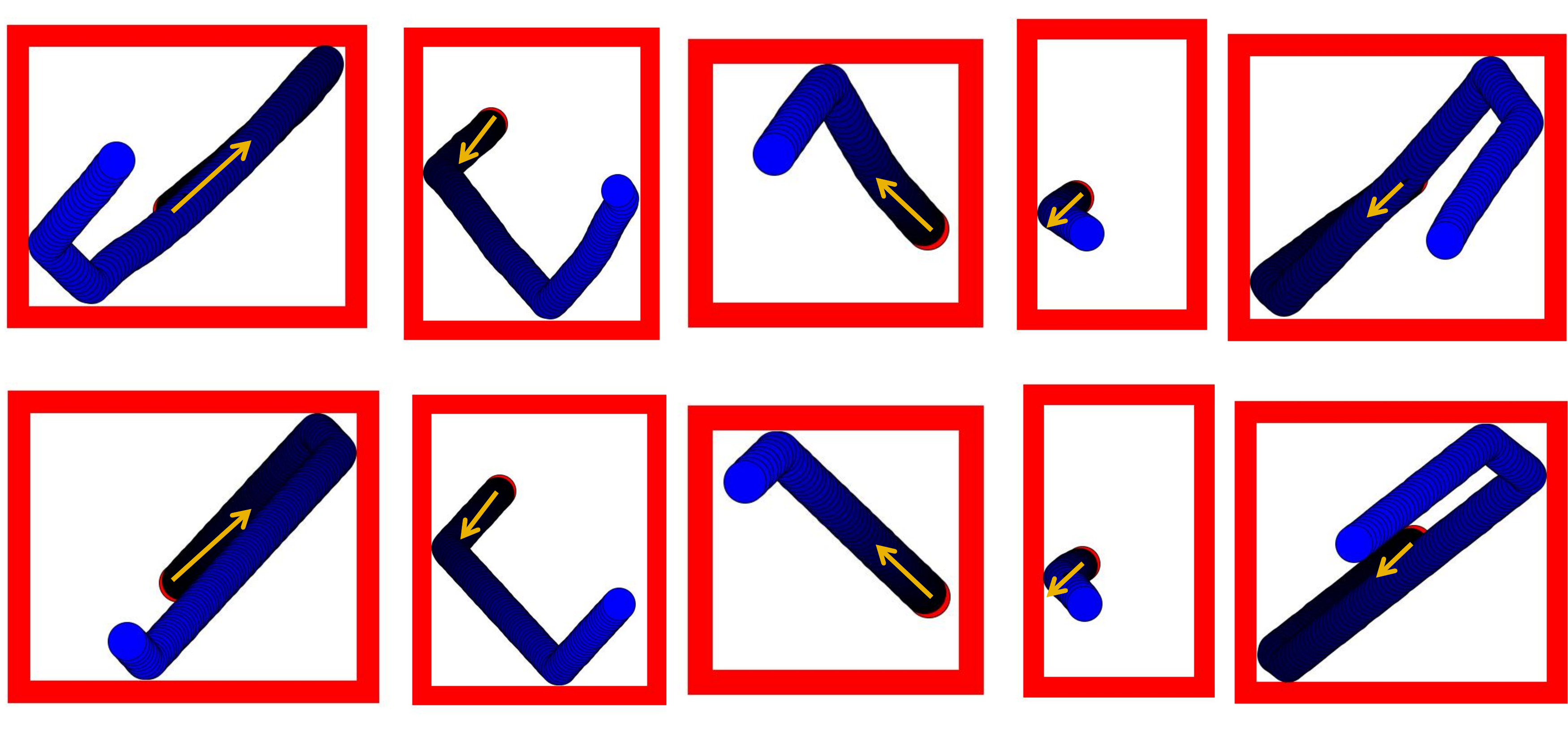}
\end{center}
\caption{Visual Imaginations generated by our model. (Top) shows the trajectory of the ball generated by our model by iteratively rendering the next world image using the predicted ball velocities and feeding this rendered image as input to the model. The imagined ball trajectory is color coded to reflect the progression of time. Change in color from dark to light blue indicates moving forward in time. (Bottom) shows the respective ground truth trajectories. Force on the ball is applied at the first time step and the force vector is shown by the orange arrow. The figure shows that our model learns the dynamics of balls and collisions and is successfully able to image collision events.}
\label{fig:hallucinations}
\end{figure}

Some examples of visual imaginations by our model are shown in figure \ref{fig:hallucinations}. Our model learns to anticipate collisions early in time. Predicting trajectory reversals at collisions is not possible using methods Kalman filter based methods that are extensively used in object tracking\citep{Welch:1995:IKF:897831}. Comparison with ground truth trajectories reveals that our models are not perfect and in some cases accumulation of errors can produce imagined trajectories that are different from the ground truth trajectories (for instance see the first column in figure \ref{fig:hallucinations}). Even in the cases when predicted trajectories do not exactly match up with the ground truth trajectories, the visual imaginations are consistent with the dynamics of balls and collisions. 

\begin{figure}[t]
\begin{center}
\includegraphics[width=0.80\linewidth]{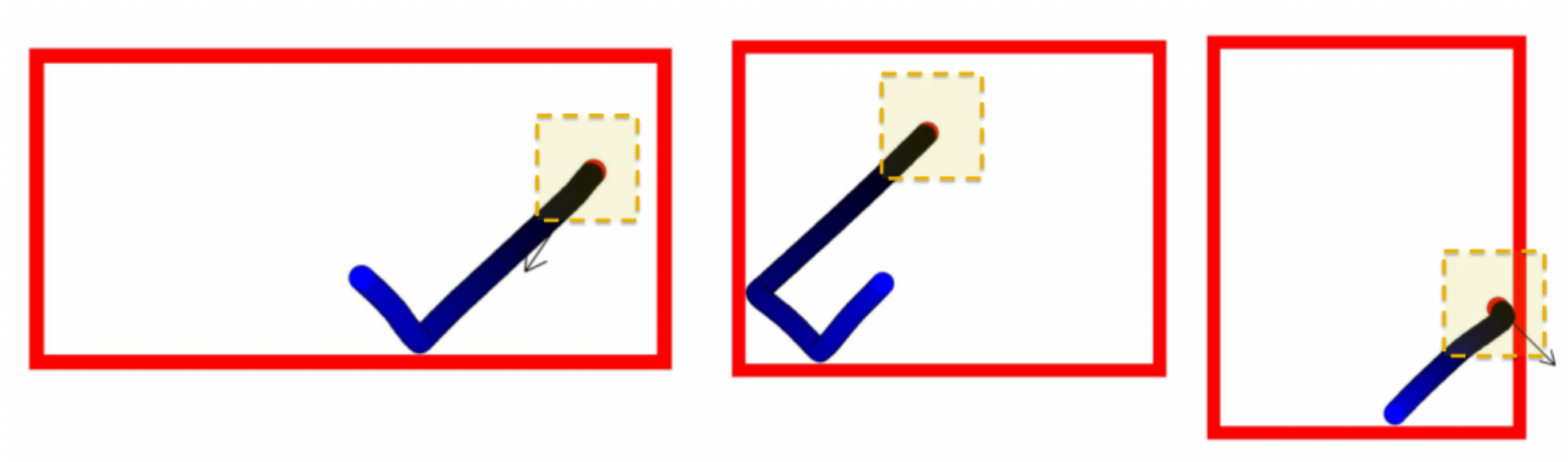}
\end{center}
\caption{Visual imaginations of our model in very large environments. Accurate predictions of the ball trajectory are made despite the fact that visual glimpses (shown in yellow) are mostly uninformative (they only contain a ball in center of white background). This is made possible by the long term memory in LSTMs. Without LSTMs, the ball exhibits non natural motion and reverses its direction in the middle of the arena.}
\label{fig:compare_lstm}
\end{figure}

Figure \ref{fig:compare_lstm} shows visual imaginations by our model in environments that are much larger than the environments used in the training set. Notice that the glimpse size is considerably smaller than the size of the environment. With glimpses of this size, visual inputs when the ball is not close to any of the walls are uninformative because such visual inputs merely comprise of a ball present in center of white background. In such scenarios, our model is able to make accurate predictions of velocity due to the long-range LSTM memory of the past. Without LSTM units, we noticed that imagined ball trajectory exhibited unexpected reversal in directions and other errors. For more examples, please see \href{https://sites.google.com/site/intuitivephysicsnips15/}{accompanying video} for imaginations in two and three ball worlds. 

\section{Using Predictive Visual Models for Action Planning}
\begin{figure}[ht]
\begin{center}
\includegraphics[width=0.75\linewidth]{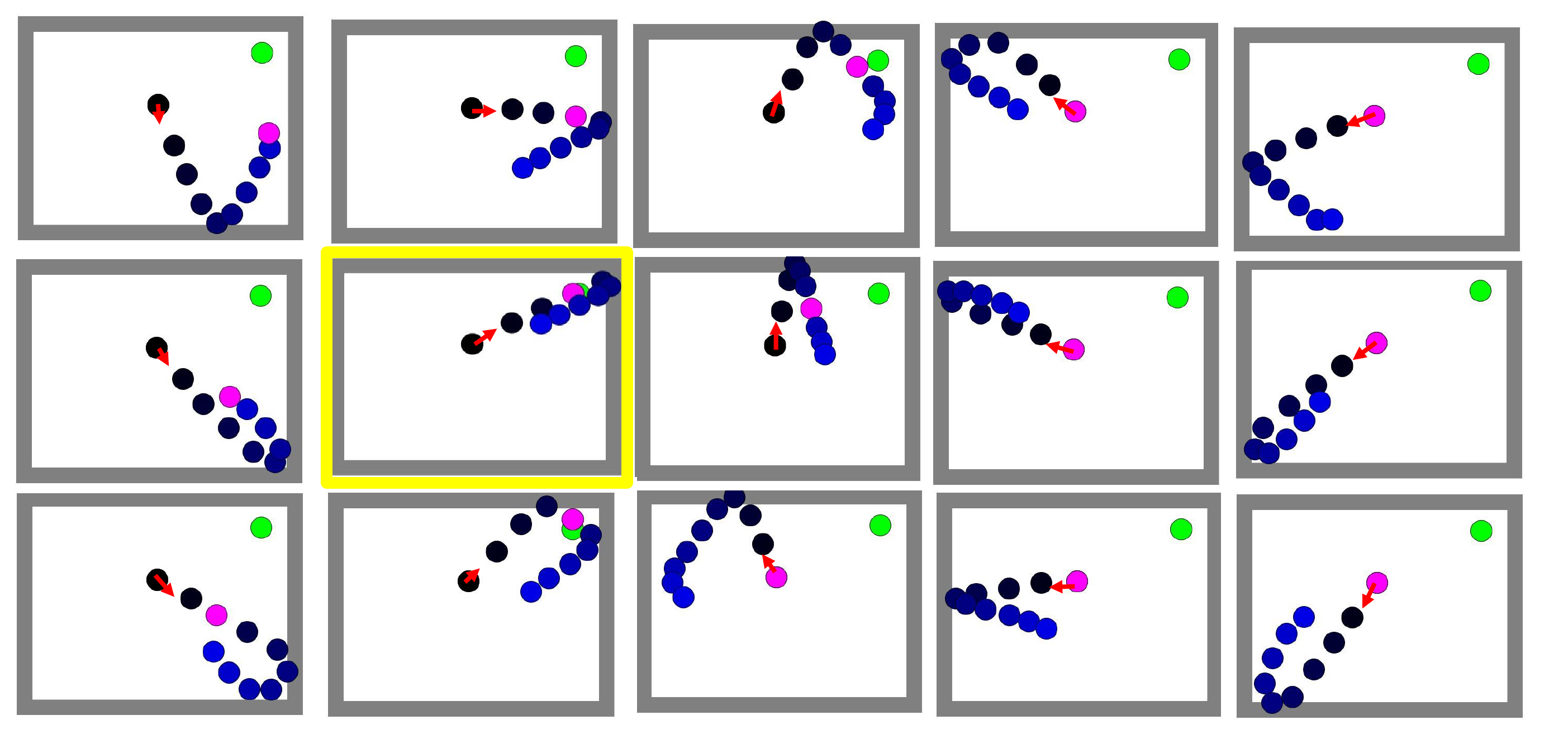}
\caption{Illustration of the method for determining the force required to push the cue ball (the ball with the arrow) to a target location (shown in green). The agent runs multiple simulations of the world (i.e. visual imagination) by applying different forces (shown by red arrows). Each grey box shows the results of simulations performed by the agent. The position of the ball closest to the target location is shown in pink in each of the simulations. The agent chooses the force that leads to a ball location that is closest to the target location (simulation in yellow box).}
\label{fig:action-plan}
\end{center}
\end{figure}

We used the learnt predictive models for planning actions to achieve goals which the agent has never received any direct supervision. We first show results on a relatively simple task of planning the force required to push the desired ball to a desired location. Next, we show results on a more challenging task of planning the force required to push the desired ball to hit a second moving ball. 

Figure \ref{fig:action-plan} illustrates the method of action planning. Given a target state, the optimal force is found by running multiple simulations (i.e. visual imaginations) of the world after applying different forces. The optimal force is the one that produces the world state that is closest to the target state. In order to verify the accuracy of this method, we use the predicted force from our model as input to our physics engine to generate its actual (rather than the imagined) outcome and compare the resulting states against the goal states. In practice, instead of exhaustively searching for all forces we use CMA-ES method \citep{hansen2001completely} for determining the optimal force. 

Table \ref{tab:action} reports the hit accuracy of our system in pushing the ball to a desired location. The hit accuracy was measured as the number of trials for which the closest point on the ball's trajectory was within $p$ pixels of the target. With an accuracy of 56\% our model is able to push the ball within 25 pixels (the size of the arena was between 300-550 pixels in size) of the target location as compared to the oracle which is successful 100\% times. The OC model significantly outperforms the FC model. The oracle was constructed by using the ground truth physics simulator for making predictions and used the same mechanism for action selection as described above. Qualitative results of our methods are best seen in the \href{https://sites.google.com/site/intuitivephysicsnips15/}{accompanying video}. We will include quantitative evaluation of more complex actions in the next revision of the paper. 

\begin{center}
\begin{table}
\begin{center}
 \scalebox{0.80}{
  \begin{tabular}{ l | c | c | c}
    Method & \multicolumn{3}{c}{Hit Accuracy} \\
    \hline
    & $<10$ pixels & $<25 $ pixels & $<50$ pixels \\ \hline
    Oracle & 95\% & 100\% & 100\% \\ \hline
    Random & 3\% & 14\% & 23\%\\ \hline \hline
    Ours (FC-Model) & 15\% & 39\% & 60\% \\\hline
    \textbf{Ours (OC-Model)} & 30\% & 56\% & 85\%\\ 
    \hline
  \end{tabular}}
  \caption{Hit accuracy of our approach for pushing the ball to a desired target location. The hit accuracy was measured as the number of trials for which the closest point on the ball's trajectory was within $p$ pixels of the target. Accuracy has been reported for p=\{10, 25, 50\} pixels. The random baseline was constructed by randomly choosing a force and the oracle used the ground truth physics simulator for selecting the optimal actions.}  
 \label{tab:action}
  \end{center}
\end{table}
\end{center}

\section{Discussion and Conclusion}
We have presented an object-centric prediction approach that exploits translation invariance in dynamics of physical systems to learn a dynamical model of the world directly from visual inputs. We show that the model generalizes to environments never encountered during training and can be used for planning actions in novel environments without the requirement of task-specific supervision.



Using our method in complex real world settings requires more nuanced mechanisms for creating visual renderings. We are investigating multiple directions like creating imaginations in a latent abstract feature space, or using visual exemplars as proxies of per frame visual  renderings. We are also exploring different mechanisms for improving predictions using error denoising and alternate loss functions. We believe that the direction of learning to predict the effect of agent's actions on the world directly from visual inputs is an important direction for enabling robots to act in previously unseen environments. Our work makes a small step in this direction.

\bibliographystyle{iclr2016_conference}
\bibliography{ballbibtex}

\begin{thebibliography}{31}
\providecommand{\natexlab}[1]{#1}
\providecommand{\url}[1]{\texttt{#1}}
\expandafter\ifx\csname urlstyle\endcsname\relax
  \providecommand{\doi}[1]{doi: #1}\else
  \providecommand{\doi}{doi: \begingroup \urlstyle{rm}\Url}\fi

\bibitem[Bhat et~al.(2002)Bhat, Seitz, and Popovic]{conf/eccv/BhatSP02}
Bhat, Kiran~S., Seitz, Steven~M., and Popovic, Jovan.
\newblock Computing the physical parameters of rigid-body motion from video.
\newblock In Heyden, Anders, Sparr, Gunnar, Nielsen, Mads, and Johansen, Peter
  (eds.), \emph{ECCV (1)}, volume 2350 of \emph{Lecture Notes in Computer
  Science}, pp.\  551--565. Springer, 2002.
\newblock ISBN 3-540-43745-2.
\newblock URL
  \url{http://dblp.uni-trier.de/db/conf/eccv/eccv2002-1.html#BhatSP02}.

\bibitem[Boots et~al.(2014)Boots, Byravan, and Fox]{DBLP:conf/icra/BootsBF14}
Boots, Byron, Byravan, Arunkumar, and Fox, Dieter.
\newblock Learning predictive models of a depth camera {\&} manipulator from
  raw execution traces.
\newblock In \emph{2014 {IEEE} International Conference on Robotics and
  Automation, {ICRA} 2014, Hong Kong, China, May 31 - June 7, 2014}, pp.\
  4021--4028, 2014.
\newblock \doi{10.1109/ICRA.2014.6907443}.
\newblock URL \url{http://dx.doi.org/10.1109/ICRA.2014.6907443}.

\bibitem[Brand et~al.(1997)Brand, Oliver, and
  Pentland]{Brand:1997:CHM:794189.794420}
Brand, M., Oliver, N., and Pentland, A.
\newblock Coupled hidden markov models for complex action recognition.
\newblock In \emph{CVPR}, 1997.

\bibitem[Brubaker et~al.(2009)Brubaker, Sigal, and Fleet]{5459407}
Brubaker, Marcus~A., Sigal, L., and Fleet, D.J.
\newblock Estimating contact dynamics.
\newblock In \emph{Computer Vision, 2009 IEEE 12th International Conference
  on}, pp.\  2389--2396, Sept 2009.
\newblock \doi{10.1109/ICCV.2009.5459407}.

\bibitem[Ghahramani \& Jordan(1997)Ghahramani and
  Jordan]{Ghahramani:1997:FHM:274158.274165}
Ghahramani, Zoubin and Jordan, Michael~I.
\newblock Factorial hidden markov models.
\newblock \emph{Mach. Learn.}, 29, 1997.

\bibitem[Hamrick et~al.(2011)Hamrick, Battaglia, and
  Tenenbaum]{hamrick2011internal}
Hamrick, Jessica, Battaglia, Peter, and Tenenbaum, Joshua~B.
\newblock Internal physics models guide probabilistic judgments about object
  dynamics.
\newblock In \emph{Proceedings of the 33rd annual conference of the cognitive
  science society}, pp.\  1545--1550. Cognitive Science Society Austin, TX,
  2011.

\bibitem[Hansen \& Ostermeier(2001)Hansen and Ostermeier]{hansen2001completely}
Hansen, Nikolaus and Ostermeier, Andreas.
\newblock Completely derandomized self-adaptation in evolution strategies.
\newblock \emph{Evolutionary computation}, 9\penalty0 (2):\penalty0 159--195,
  2001.

\bibitem[Haruno et~al.(2001)Haruno, Wolpert, and Kawato]{haruno2001mosaic}
Haruno, Masahiko, Wolpert, David~H, and Kawato, Mitsuo.
\newblock Mosaic model for sensorimotor learning and control.
\newblock \emph{Neural computation}, 13\penalty0 (10):\penalty0 2201--2220,
  2001.

\bibitem[Hofsten \& Rosander(1997)Hofsten and Rosander]{Hofsten19971799}
Hofsten, Claes~Von and Rosander, Kerstin.
\newblock Development of smooth pursuit tracking in young infants.
\newblock \emph{Vision Research}, 37\penalty0 (13):\penalty0 1799 -- 1810,
  1997.
\newblock ISSN 0042-6989.
\newblock \doi{http://dx.doi.org/10.1016/S0042-6989(96)00332-X}.
\newblock URL
  \url{http://www.sciencedirect.com/science/article/pii/S004269899600332X}.

\bibitem[Jordan \& Rumelhart(1992)Jordan and Rumelhart]{jordan1992forward}
Jordan, Michael~I and Rumelhart, David~E.
\newblock Forward models: Supervised learning with a distal teacher.
\newblock \emph{Cognitive science}, 16\penalty0 (3):\penalty0 307--354, 1992.

\bibitem[Kietzmann \& Riedmiller(2009)Kietzmann and
  Riedmiller]{DBLP:conf/icmla/KietzmannR09}
Kietzmann, Tim~C. and Riedmiller, Martin~A.
\newblock The neuro slot car racer: Reinforcement learning in a real world
  setting.
\newblock In \emph{International Conference on Machine Learning and
  Applications, {ICMLA} 2009, Miami Beach, Florida, USA, December 13-15, 2009},
  pp.\  311--316, 2009.
\newblock \doi{10.1109/ICMLA.2009.15}.
\newblock URL \url{http://dx.doi.org/10.1109/ICMLA.2009.15}.

\bibitem[Krizhevsky et~al.(2012)Krizhevsky, Sutskever, and
  Hinton]{NIPS2012_0534}
Krizhevsky, Alex, Sutskever, Ilya, and Hinton, Geoffrey~E.
\newblock Imagenet classification with deep convolutional neural networks.
\newblock In \emph{NIPS}. 2012.

\bibitem[Kyriazis et~al.(2011)Kyriazis, Oikonomidis, and
  Argyros]{DBLP:conf/bmvc/KyriazisOA11}
Kyriazis, Nikolaos, Oikonomidis, Iason, and Argyros, Antonis~A.
\newblock Binding computer vision to physics based simulation: The case study
  of a bouncing ball.
\newblock In \emph{British Machine Vision Conference, {BMVC} 2011, Dundee, UK,
  August 29 - September 2, 2011. Proceedings}, pp.\  1--11, 2011.
\newblock \doi{10.5244/C.25.43}.
\newblock URL \url{http://dx.doi.org/10.5244/C.25.43}.

\bibitem[Lange et~al.(2012)Lange, Riedmiller, and
  Voigtl{\"{a}}nder]{DBLP:conf/ijcnn/LangeRV12}
Lange, Sascha, Riedmiller, Martin~A., and Voigtl{\"{a}}nder, Arne.
\newblock Autonomous reinforcement learning on raw visual input data in a real
  world application.
\newblock In \emph{The 2012 International Joint Conference on Neural Networks
  (IJCNN), Brisbane, Australia, June 10-15, 2012}, pp.\  1--8, 2012.
\newblock \doi{10.1109/IJCNN.2012.6252823}.
\newblock URL \url{http://dx.doi.org/10.1109/IJCNN.2012.6252823}.

\bibitem[Levine et~al.(2015)Levine, Finn, Darrell, and
  Abbeel]{DBLP:journals/corr/LevineFDA15}
Levine, Sergey, Finn, Chelsea, Darrell, Trevor, and Abbeel, Pieter.
\newblock End-to-end training of deep visuomotor policies.
\newblock \emph{CoRR}, abs/1504.00702, 2015.
\newblock URL \url{http://arxiv.org/abs/1504.00702}.

\bibitem[Lillicrap et~al.(2015)Lillicrap, Hunt, Pritzel, Heess, Erez, Tassa,
  Silver, and Wierstra]{DeepMindContinuous}
Lillicrap, Timothy~P., Hunt, Jonathan~J., Pritzel, Alexander, Heess, Nicolas,
  Erez, Tom, Tassa, Yuval, Silver, David, and Wierstra, Daan.
\newblock Continuous control with deep reinforcement learning.
\newblock \emph{CoRR}, abs/1509.02971, 2015.

\bibitem[Mayne(2014)]{Mayne20142967}
Mayne, David~Q.
\newblock Model predictive control: Recent developments and future promise.
\newblock \emph{Automatica}, 50\penalty0 (12):\penalty0 2967 -- 2986, 2014.
\newblock ISSN 0005-1098.
\newblock \doi{http://dx.doi.org/10.1016/j.automatica.2014.10.128}.
\newblock URL
  \url{http://www.sciencedirect.com/science/article/pii/S0005109814005160}.

\bibitem[Michalski et~al.(2014)Michalski, Memisevic, and
  Konda]{DBLP:conf/nips/MichalskiMK14}
Michalski, Vincent, Memisevic, Roland, and Konda, Kishore~Reddy.
\newblock Modeling deep temporal dependencies with recurrent grammar cells"".
\newblock In \emph{Advances in Neural Information Processing Systems 27: Annual
  Conference on Neural Information Processing Systems 2014, December 8-13 2014,
  Montreal, Quebec, Canada}, pp.\  1925--1933, 2014.
\newblock URL
  \url{http://papers.nips.cc/paper/5549-modeling-deep-temporal-dependencies-with-recurrent-grammar-cells}.

\bibitem[Mnih et~al.(2013)Mnih, Kavukcuoglu, Silver, Graves, Antonoglou,
  Wierstra, and Riedmiller]{DBLP:journals/corr/MnihKSGAWR13}
Mnih, Volodymyr, Kavukcuoglu, Koray, Silver, David, Graves, Alex, Antonoglou,
  Ioannis, Wierstra, Daan, and Riedmiller, Martin~A.
\newblock Playing atari with deep reinforcement learning.
\newblock \emph{CoRR}, abs/1312.5602, 2013.
\newblock URL \url{http://arxiv.org/abs/1312.5602}.

\bibitem[Mottaghi et~al.(2015)Mottaghi, Bagherinezhad, Rastegari, and
  Farhadi]{DBLP:journals/corr/MottaghiBRF15}
Mottaghi, Roozbeh, Bagherinezhad, Hessam, Rastegari, Mohammad, and Farhadi,
  Ali.
\newblock Newtonian image understanding: Unfolding the dynamics of objects in
  static images.
\newblock \emph{CoRR}, abs/1511.04048, 2015.
\newblock URL \url{http://arxiv.org/abs/1511.04048}.

\bibitem[Oh et~al.(2015)Oh, Guo, Lee, Lewis, and Singh]{actionconditioned}
Oh, Junhyuk, Guo, Xiaoxiao, Lee, Honglak, Lewis, Richard, and Singh, Satinder.
\newblock Action-conditional video prediction using deep networks in atari
  games.
\newblock \emph{arXiv preprint arXiv:1507.08750}, 2015.

\bibitem[Salzmann \& Urtasun(2011)Salzmann and Urtasun]{conf/iccv/SalzmannU11}
Salzmann, Mathieu and Urtasun, Raquel.
\newblock Physically-based motion models for 3d tracking: A convex formulation.
\newblock In Metaxas, Dimitris~N., Quan, Long, Sanfeliu, Alberto, and Gool, Luc
  J.~Van (eds.), \emph{ICCV}, pp.\  2064--2071. IEEE, 2011.
\newblock ISBN 978-1-4577-1101-5.
\newblock URL
  \url{http://dblp.uni-trier.de/db/conf/iccv/iccv2011.html#SalzmannU11}.

\bibitem[Sutskever et~al.(2008)Sutskever, Hinton, and Taylor]{sht-rtrbm-08}
Sutskever, Ilya, Hinton, Geoffrey~E., and Taylor, Graham~W.
\newblock The recurrent temporal restricted boltzmann machine.
\newblock In \emph{NIPS}, 2008.

\bibitem[Todorov \& Ghahramani(2003)Todorov and
  Ghahramani]{todorov2003unsupervised}
Todorov, Emanuel and Ghahramani, Zoubin.
\newblock Unsupervised learning of sensory-motor primitives.
\newblock In \emph{Engineering in Medicine and Biology Society, 2003.
  Proceedings of the 25th Annual International Conference of the IEEE},
  volume~2, pp.\  1750--1753. IEEE, 2003.

\bibitem[Urtasun et~al.(2006)Urtasun, Fleet, and
  Fua]{Urtasun:2006:PTG:1153170.1153448}
Urtasun, Raquel, Fleet, David~J., and Fua, Pascal.
\newblock 3d people tracking with gaussian process dynamical models.
\newblock In \emph{CVPR}, 2006.

\bibitem[Wahlstr{\"{o}}m et~al.(2015)Wahlstr{\"{o}}m, Sch{\"{o}}n, and
  Deisenroth]{DBLP:journals/corr/WahlstromSD15}
Wahlstr{\"{o}}m, Niklas, Sch{\"{o}}n, Thomas~B., and Deisenroth, Marc~Peter.
\newblock From pixels to torques: Policy learning with deep dynamical models.
\newblock \emph{CoRR}, abs/1502.02251, 2015.
\newblock URL \url{http://arxiv.org/abs/1502.02251}.

\bibitem[Watter et~al.(2015)Watter, Springenberg, Boedecker, and
  Riedmiller]{DBLP:journals/corr/WatterSBR15}
Watter, Manuel, Springenberg, Jost~Tobias, Boedecker, Joschka, and Riedmiller,
  Martin~A.
\newblock Embed to control: {A} locally linear latent dynamics model for
  control from raw images.
\newblock \emph{CoRR}, abs/1506.07365, 2015.
\newblock URL \url{http://arxiv.org/abs/1506.07365}.

\bibitem[Welch \& Bishop(1995)Welch and Bishop]{Welch:1995:IKF:897831}
Welch, Greg and Bishop, Gary.
\newblock An introduction to the kalman filter.
\newblock Technical report, Chapel Hill, NC, USA, 1995.

\bibitem[Weng et~al.(2006)Weng, Kuo, and Tu]{Weng:2006:VOT:1223195.1223208}
Weng, Shiuh-Ku, Kuo, Chung-Ming, and Tu, Shu-Kang.
\newblock Video object tracking using adaptive kalman filter.
\newblock \emph{J. Vis. Comun. Image Represent.}, 17\penalty0 (6):\penalty0
  1190--1208, 2006.
\newblock ISSN 1047-3203.
\newblock \doi{10.1016/j.jvcir.2006.03.004}.
\newblock URL \url{http://dx.doi.org/10.1016/j.jvcir.2006.03.004}.

\bibitem[Wolpert et~al.(1995)Wolpert, Ghahramani, and
  Jordan]{wolpert1995internal}
Wolpert, Daniel~M, Ghahramani, Zoubin, and Jordan, Michael~I.
\newblock An internal model for sensorimotor integration.
\newblock \emph{Science-AAAS-Weekly Paper Edition}, 269\penalty0
  (5232):\penalty0 1880--1882, 1995.

\bibitem[Wu et~al.(2015)Wu, Yildirim, Lim, Freeman, and
  Tenenbaum]{NIPS2015_5780}
Wu, Jiajun, Yildirim, Ilker, Lim, Joseph~J, Freeman, Bill, and Tenenbaum, Josh.
\newblock Galileo: Perceiving physical object properties by integrating a
  physics engine with deep learning.
\newblock In Cortes, C., Lawrence, N.D., Lee, D.D., Sugiyama, M., Garnett, R.,
  and Garnett, R. (eds.), \emph{Advances in Neural Information Processing
  Systems 28}, pp.\  127--135. Curran Associates, Inc., 2015.
\newblock URL
  \url{http://papers.nips.cc/paper/5780-galileo-perceiving-physical-object-properties-by-integrating-a-physics-engine-with-deep-learning.pdf}.

\end{thebibliography}

\end{document}